\documentclass[article]{clv3}
\usepackage{natbib}
\usepackage{graphicx}
\graphicspath{{figures/}}
\usepackage{CJKutf8}
\title{Universal Syntactic Structures: Modeling Syntax for Various Natural Languages}
\author[1]{Min K. Kim \\ minerica@uw.edu }
\author[2]{Hafu Takero \\ takero@uw.edu}
\author[3]{Sara Fedovik \\ fdvk@uw.edu}
\runningtitle{Universal Syntactic Structures}
\runningauthor{Kim}
\begin{document}
\begin{CJK}{UTF8}{mj}
\maketitle
\begin{abstract}
We aim to provide an explanation for how the human brain might connect words for sentence formation. A novel approach to modeling syntactic representation is introduced, potentially showing the existence of universal syntactic structures for all natural languages. As the discovery of DNA's double helix structure shed light on the inner workings of genetics, we wish to introduce a basic understanding of how language might work in the human brain. It could be the brain's way of encoding and decoding knowledge. It also brings some insight into theories in linguistics, psychology, and cognitive science.
\par
After looking into the logic behind universal syntactic structures and the methodology of the modeling technique, we attempt to analyze corpora that showcase universality in the language process of different natural languages such as English and Korean. Lastly, we discuss the critical period hypothesis, universal grammar, and a few other assertions on language for the purpose of advancing our understanding of the human brain.
\end{abstract}
\bigskip
\section{Introduction}
Having the ability to create and understand complex thoughts is what fundamentally separates humans from other mammals. These discrete thoughts are often referred to as \emph{sentences}. Language is the system of vocabulary and syntax that enables humans to generate and process structural connections of words. In essence, it consists of an encoding and decoding mechanism in the form of natural languages. The language process produces semantics or meaning, which gives us an ability to understand abstract or highly complex concepts. Although the word is usually considered to be a form of communication, it is better to define \emph{language} as a code or cipher for generating discrete thoughts. Because we happen to be a social species, language is also utilized for communication as well.
\par
But how does language actually work? More specifically, how does the human brain connect words to form sentences? Despite an enormous amount of effort by scientists and engineers, we have yet reached a plateau where we can undoubtedly declare machines are our equals when it comes to natural language processing (NLP). The best chess players and go players in the world are no longer humans. However, the best translators of natural languages are still humans and not machines. Machine translation (MT) has not reached a level of human translators. But if we were to apply the same process used by the language faculty, it will no longer be such an impossible task for machines to process language at the same level as humans in terms of quality and performance. Most likely, the machine will eventually exceed our linguistic capabilities in both speed and precision. Computers have already become faster and better at playing board games, making mathematical calculations, creating images, etc. It is likely only a matter of time before they conquer NLP as well. But this feat would require a comprehensive understanding of what language is and how it works. Machines are not built to compute numbers by guessing the right outcome based on data and training. Having a simple, logical process with transistors and logic gates is clearly a better approach. Similarly, finding a more optimal process for language is needed for future language models.
\par
In this article, we propose a novel modeling technique for representing syntax. This method appears to exhibit a level of performance that is equal to humans. We suspect it could be an indication of a discovery, potentially revealing how the brain processes words to create discrete thoughts. In addition, we attempt to provide a logical explanation for why it could be the biological NLP system for the human brain.
\par
Words in a sentence can be arranged in different ways for different languages, and yet, they still convey the same meaning. Variation in syntax has served as an indication that natural languages are intrinsically heterogeneous rather than homogeneous. Noam Chomsky, on the other hand, has been a proponent of universal grammar (UG), a theory that all natural languages share essentially the same grammar or syntax at a hidden level (Barman 2012). If humans are equipped with the identical language faculty from birth, then first-language (L1) acquisition takes place in children the exact same way regardless of which natural languages they are exposed to. A Jamaican child being raised in London learns English just like any other children in the same region. A child with Polish parents will learn to speak Filipino fluently if she grows up in the Philippines and not Poland. However, this seems to be strictly a trait of Homo sapiens. Numerous scientists in the past have made attempts to instill language into chimpanzees and gorillas, but not a single mammal could learn to speak or sign like humans do. This suggests animals are not equipped with the language faculty, which would be required for combining words together to create discrete sentences and thoughts.
\par
Linguists have used syntax trees or parse trees to represent sentences visually. Syntax trees are used to illustrate sentences as hierarchical structures with lexical categories describing the type of each word. The syntactic structure of a sentence can shape its meaning like the lexical semantics of individual words. Because of this reason, there might not be any difference between \emph{semantics} and \emph{syntax} since both concepts are all part of mental representations (Chomsky 2000). However, this conjecture appears to be false as two sentences in two different languages can convey the exact same meaning while having completely different syntactic structures. Furthermore, hierarchical representations of syntax imply there exists an unknown property of sentence formation, which causes words to be linked together grammatically to create sentences. No one has been able to demonstrate how such a process can take place in the brain to generate discrete thoughts almost instantaneously. The swiftness of sentence formation suggests language utilization is more of a simple process than a complex one.
\par
Practical applications of syntax trees, so far, have been very limited in scope and use. The descriptive nature of syntax trees makes them useful for sentence analysis but not necessarily for sentence formation or translation. Recently, the development of Universal Dependencies (UD) have become more of a practical solution for real-world applications. Because UD can show how content words such as nouns and verbs in the same sentence are related to each other, it can be quite useful for sentence parsing (Kondratyuk and Straka 2019). The deep-learning paradigm has also drastically improved the performance of NLP. Language models such as BERT (Bidirectional Encoder Representations from Transformers) and GPT-4 (Generative Pretrained Transformer 4) have the ability to generate sentences and paragraphs that may be indistinguishable from the ones created by humans. The combination of having large data of corpus and the recent advancement of NLP research has significantly improved the quality of machine translation over the last decade. Neural machine translation (NMT) has shown improvements to previous approaches such as statistical machine translation (SMT) by training artificial neural networks to yield better translation results (Bahdanau, Cho, and Bengio 2015). However, the new approach still remains as a probabilistic model. This means getting near 100\% accurate results in translation is highly unrealistic since the output is always more or less a prediction. Clearly, this cannot be the optimal solution for MT since humans do not demonstrate similar problems and have better performance of translation. Although NMT seems to be effective, it is not without its own set of limitations (Castilho et al. 2017). Understanding how the human brain processes language can potentially eliminate these limitations due to the fact that implementing the same system will likely yield comparable outcomes. In theory, it may even be difficult to distinguish humans from machines when it comes to language processing.
\section{Sentence Types and Word Orders}
Sentences can be classified as one of the four types; declarative (statements), interrogative (questions), imperative (commands and requests), and exclamatory (emotional expressions). We exclude commonly used phrases such as \emph{No problem}, \emph{Ouch}, and \emph{Go figure} as sentences in our analysis because they may not contain proper syntax or meaningful grammatical elements. Therefore, we view a grammatical sentence as a syntactic construction of words. This means a sentence should be able to be tokenized into separate, independent words in order for them to be connected to each other. A grammatical sentence has a subject and a verb. It is common in some languages to omit the subject and not explicitly state it. But this would require the recipient of the sentence–as either the listener or the reader–to be expected to know about the subject. For example, an imperative sentence such as \emph{Please, don't change the channel} does not explicitly state what the subject is. But it is obvious to the listener that the subject is a second-person pronoun, referring to the listener. A sentence can also have one or several objects that modify the verb. The order the subject, verb, object are positioned determines the word order of the sentence. In English, declarative sentences usually have the word order of subject-verb-object or SVO. The other five word orders are subject-object-verb (SOV), verb-subject-object (VSO), verb-object-subject (VOS), object-subject-verb (OSV), and object-verb-subject (OVS).
\par
Three different tokens yield six permutations or arrangements; \begin{math} 3! = 3 \times 2 \times 1 \end{math}. For instance, three numbers–one, two, and three–can be arranged as the following: 123, 132, 213, 231, 312, and 321. Similarly, splitting a sentence into three components–subject (S), verb (V), and object (O)–yields the same number of permutations. Therefore, once again, the six possible word orders are SVO, SOV, VSO, VOS, OSV, and OVS. The main word order used in English is SVO (e.g. \emph{Mary loves chocolate}). Other languages might use different word orders such as SOV to create declarative sentences (e.g. \emph{Mary chocolate loves}). All the six word orders are used by different natural languages, especially SOV and SVO (Carnie 2002). These word orders can be unified as a single structure by looping the three components together.
\clearpage
\begin{figure}[ht]
    \centering
    \includegraphics[width=0.75\textwidth]{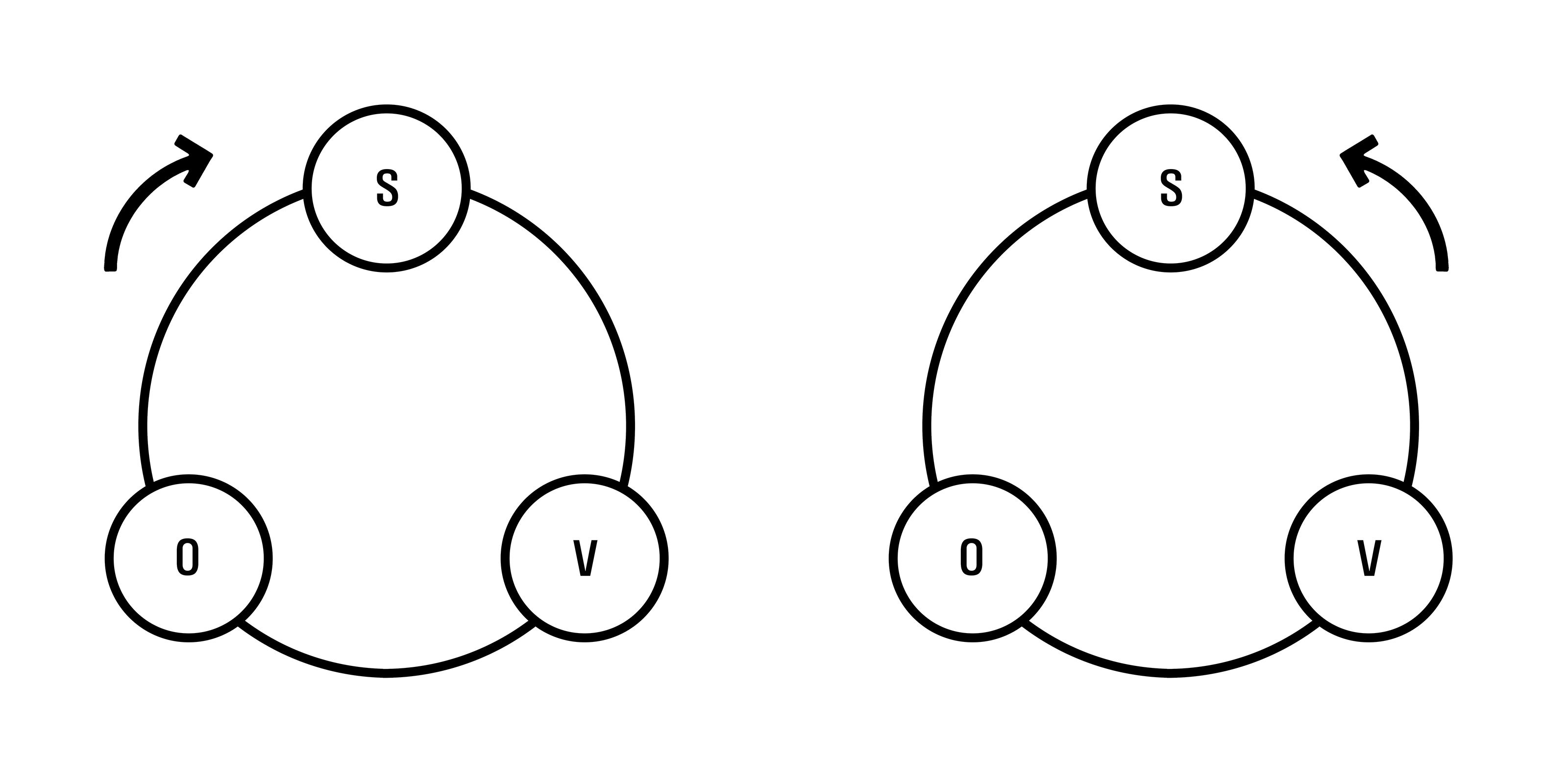}
    \caption{All the six word orders belong to one structure by moving the three components either clockwise (SVO, VOS, OSV) or counterclockwise (SOV, OVS, VSO).}
    \label{fig:fig01}
\end{figure}
\par
The clockwise rotation gives rise to three of the six word orders whereas the counterclockwise rotation is responsible for the remaining three. If a declarative sentence in English (SVO) is considered to move clockwise, then the equivalent sentence in Japanese (SOV) would move counterclockwise and vice versa. This is called the \emph{direction of flow}. The word order is not set per natural language since one language can make use of different word orders. However, a sentence in any language must belong to one particular word order as it cannot be SOV and something else at the same time.
\section{The Synapper}
A loop can be created by connecting the first and the last constituents of a sentence for any language in any given word order. But sentences might require some words to be linked in two or more given directions or dimensions. (We hereby reserve the word \emph{direction} for only describing the entire process of parsing words in a loop as clockwise or counterclockwise to avoid any confusion. The word \emph{dimension} is used for describing individual connections of words.) If a word such as \emph{blue} depends on the presence of another word such as \emph{bird}, then the dependent word is linked to only one word and is not a part of the main loop. We define this approach of symbolic representation as the \emph{synapper}. The synapper is a mechanism that utilizes looping in multiple dimensions in order to connect tokens such as words. In linguistics, it attempts to represent syntactic structures. By modeling the synapper, translations of even complex sentences in different word orders can be merged to form a single, unifying structure.
\par
We use the following declarative sentence as an example to investigate whether the syntactic structure is uniform for English (SVO), French (SVO), Japanese (SOV), Uzbek (SOV), and Welsh (VSO):
\begin{center}
Jane has a very fast brown horse.
\end{center}
\par
It has the default subject-verb-object (SVO) arrangement in English, where the predicate is composed of verb + determiner + adverb + adjective + adjective + noun:
\begin{center}
V + DET + ADV + ADJ + ADJ + N
\end{center}
\par
By arranging these words in more than one dimension, we can create the synapper model of the sentence. The words that belong to the main loop or circuit are called \emph{nodes}. Any word that is connected to a node, modifying it from a different dimension, is called a \emph{branch}. A \emph{constituent} is defined as a node with its branches.
\begin{figure}[ht]
    \centering
    \includegraphics[width=0.75\textwidth]{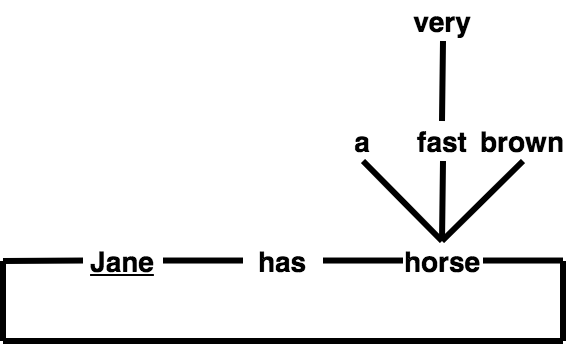}
    \caption{The starting constituent for English is underlined (Jane). In SVO languages, the sentence is read clockwise starting with the subject. The branch words that are connected to the node horse are read with the far-left word first (a, very, fast, brown). In some languages like French and Spanish, some branch words are supposed to be read after the node (a, horse, brown, very, fast).}
    \label{fig:fig02}
\end{figure}
\par
Here is the breakdown of the way the words in the sentence are ordered in each language:
\begin{itemize}
\setlength{\itemindent}{0.1in}
\item English: Jane has a very fast brown horse.
\item French: Jane has a horse brown very fast. (Jane a un cheval brun très rapide.)
\item Japanese: Jane very fast brown horse has. (ジェーンはとても早い茶色の馬を持っている。)
\item Uzbek: Jane very fast brown horse has. (Janeda bir juda tez jigarrang ot bor.)
\item Welsh: Has Jane horse brown very fast. (Mae gan Jane geffyl brown cyflym iawn.)
\end{itemize}
\par
Because SOV and VSO have the same direction of flow (e.g. counterclockwise), the sentence in Japanese, Uzbek, and Welsh should move in the same direction. The only difference is Japanese and Uzbek start with the subject \emph{Jane} whereas Welsh starts with the verb \emph{has}. For English and French, the sentence is read in the opposite direction (e.g. clockwise) since SVO belongs to the other group with VOS and OSV.
\par
This means we can take the synapper model in Figure \ref{fig:fig02} and derive the correct translation for each language. In other words, a single syntactic structure has all the sufficient information for expressing the same thought in any natural language as long as the word order and the direction of flow are given. For instance, this structure can yield the following sentence by traveling counterclockwise starting with the subject:
\begin{center}
Jane very fast brown horse has.
\end{center}
\par
Now we can simply replace the English words with words in Uzbek and then morphemes can be added, changed, or removed such as the determiner \emph{a} based on the target language's grammar. The result is \emph{Janeda bir juda tez jigarrang ot bor}, which is the correct Uzbek translation.
\par
Creating the synapper models of interrogative sentences requires a few more steps. Languages like English switch positions of the subject noun phrase (NP) and the main verb phrase (VP) for turning a declarative sentence into a question. (VP in this context does not include anything else such as an adjective phrase.) However, this is not the case for many other languages. They use verb conjugations or other methods to create interrogative sentences. If the function of language is to create thoughts, then the declarative form of a sentence should become the default form. That means turning a declarative sentence into an interrogative style requires additional rules. These rules differ from language to language. So interrogative sentences must resort back to their declarative forms for their synapper models to work for other languages.
\par
Here is an interrogative sentence in English:
\begin{center}
Why is Tim going to the hospital?
\end{center}
\par
The sentence can become declarative by removing the word \emph{why} from the sentence and then changing the word order to SVO:
\begin{center}
Tim is going to the hospital.
\end{center}
\begin{figure}[ht]
    \centering
    \includegraphics[width=0.75\textwidth]{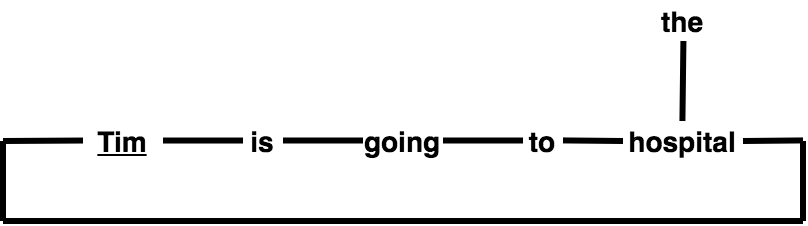}
    \caption{For SOV languages, the sentence is read counterclockwise starting with Tim (Tim, the, hospital, to, going, is).}
    \label{fig:fig03}
\end{figure}
\par
To add the word \emph{why}, different rules have to be applied. For English and French, the word is placed in the beginning of the sentence and then NP and VP are switched. In Japanese and Uzbek, the word is placed before \emph{Tim}. In Welsh, it is put in the beginning of the sentence without moving NP and VP. Therefore, the interrogative forms become as follows:
\begin{itemize}
\setlength{\itemindent}{0.1in}
\item English: Why is Tim going to the hospital?
\item French: Why is Tim going to the hospital? (Pourquoi Tim va-t-il à l'hôpital?)
\item Japanese: Why Tim the hospital to going is? (なぜティムは病院に行っているのですか？)
\item Uzbek: Why Tim the hospital to going is? (Nega Tim kasalxonaga ketayapti?)
\item Welsh: Why is Tim going to the hospital? (Pam mae Tim yn mynd i'r ysbyty?)
\end{itemize}
\par
Since interrogative sentences are essentially modified versions of declarative sentences, their grammatical rules are not necessarily identical between languages. If different languages have different rules of grammar to create interrogative sentences, then the same rules must be implemented to synapper modeling accordingly one by one.
\section{Recursion}
One of the properties of language is its ability to be recursive. A recursive sentence can be made by adding phrases like \emph{I think} or \emph{It is true that}. Recursion enables varying degrees of complexity in sentences and thoughts. To model recursion in declarative sentences, some constituents have to be embedded or layered inside the main circuit. The following is a recursive sentence:
\begin{center}
The fact that Colette was Willy was a big secret.
\end{center}
\par
The first six words make up the subject noun phrase of the sentence. Because recursion is applied twice, a loop can be formed to \emph{Colette was Willy} and then it can be looped again with the first three words of the sentence.
\begin{figure}[ht]
    \centering
    \includegraphics[width=0.75\textwidth]{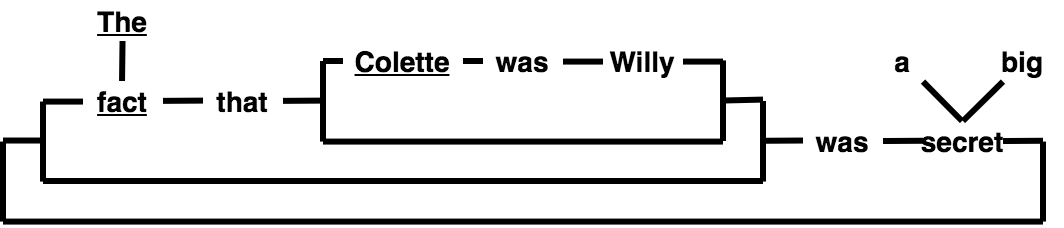}
    \caption{The subject phrase exists in multiple loops or layers. They all have the same direction of flow (either clockwise or counterclockwise) for the given sentence.}
    \label{fig:fig04}
\end{figure}
\par
The recursive inner layers travel in the same direction as the main loop, being consistent with the sentence's word order. Here is the correct arrangement in each language:
\begin{itemize}
\setlength{\itemindent}{0.1in}
\item English: The fact that Colette was Willy was a big secret.
\item French: The fact that Colette was Willy was a big secret. (Le fait que Colette soit Willy était un grand secret.)
\item Japanese: Colette Willy was that fact a big secret was. (コレットがウィリーだった事実は大きな秘密だった。)
\item Uzbek: Colette Willy was that fact a big secret was. (Colettening Willy ekanligi fakti katta sir edi.)
\item Welsh: Was the fact that Colette was Willy a big secret. (Roedd y ffaith mai Colette oedd Willy yn gyfrinach fawr.)
\end{itemize}
\par
Although one syntactic structure accurately represents the sentence in all five languages, the starting point of the sentence can be different. The first word in Japanese and Uzbek is \emph{Colette} whereas \emph{the fact} are the first two words for English and French. In Welsh, the first constituent is \emph{was} since Welsh is a VSO language. However, the direction of flow for the Welsh sentence is different from Japanese's and Uzbek's. In Figure \ref{fig:fig04}, the sentence should be read clockwise for English and French and counterclockwise for the other three languages. But the Welsh translation behaves as if it is not actually a VSO sentence. Instead, the word order appears to be the same as English, which is SVO. The only difference is the verb is placed at the beginning of the sentence for Welsh. This phenomenon can be observed in Figure \ref{fig:fig03} as well. If Welsh is truly a VSO language, then the correct order of translation should be \emph{Is Tim the hospital to going}. This would match the direction of flow of Japanese and Uzbek as it should. But the correct translation in Welsh is \emph{Is Tim going to the hospital}. This is no different from the original sentence in English except for placement of the verb. Thus, based on the evidence, we find that Welsh's actual word order is not VSO. It appears to be VSO strictly because the verb is placed before the subject. However, it cannot be a VSO language since the direction of flow matches that of SVO. So Welsh's real word order appears to be SVO/V1. V1 or \emph{verb-initial} indicates the verb must be placed before the subject regardless of the word order. Some languages have a V2 word order or \emph{verb-second} such as German, Norwegian, and Swedish. When there is a verb movement within a sentence, the verb might have to be moved back to its original position in order to match the direction of flow with the rest of the sentence. Even if a language uses V1 or V2, it does not change the fact that a particular sentence must belong to one of the six word orders (e.g. SVO/V1 = SOV). Korean and Slavic languages are considered by some as "free word order" languages since they allow certain expressions to exist in different word orders. But this still does not change the "one word order per sentence" rule.
\par
The sentence \emph{The fact that Colette was Willy was a big secret} contains ten words. We can calculate the probability of the words being placed in their correct positions in the translation of a language with a different word order as follows where \begin{math} n - x \end{math} is greater than one:
\begin{center}
\begin{math}
{\tiny P = \left( \frac{1}{n} \right) \times \left( \frac{1}{n - 1} \right) \times \left( \frac{1}{n - 2} \right) \times \left( \frac{1}{n - 3} \right) \times \cdots }
\end{math}
\end{center}
\par
Since the total word count (n) is ten, the formula becomes:
\begin{center}
\begin{math}
{\tiny P = \left( \frac{1}{10} \right) \times \left( \frac{1}{9} \right) \times \left( \frac{1}{8} \right) \times \left( \frac{1}{7} \right) \times \cdots = 0.00000027}
\end{math}
\end{center}
\par
In other words, the probability of one syntactic structure producing the correct word order for the Korean translation from an English sentence as a coincidence is less than 0.00003\%. If the number of words in a sentence rises to 15, then the odds become less than 0.00000000008\%. We view this as a strong indication that the likelihood of synapper models representing the correct syntactic structure for different natural languages, due to chance, is extremely unlikely.
\section{Ambiguity}
The concept of ambiguity raises an interesting question regarding whether the meaning of a sentence is actually morphed by its structure. An English speaker can easily tell the difference of a phrase \emph{although he knew I told him} between \emph{He was surprised, although he knew I told him} and \emph{He was surprised. Although he knew, I told him}. In the first instance, the phrase behaves as a subordinate clause. In the second sentence, \emph{although he knew} is a subordinate clause whereas \emph{I told him} is the main clause. The same words are used in the exact same order for representing two independent thoughts. Therefore, the synapper model for each expression should not be the same. The first sentence has \emph{I told him} embedded in the structure \emph{he knew X} where \emph{X} is replaced by \emph{I told him}. In the second expression, the subordinate clause \emph{although he knew} is simply inserted before the main clause, \emph{I told him}, without any embedding. As the meaning of the expression changes, the syntactic structure also changes. In other words, the meaning changes as a sentence's syntactic structure changes.
\par
We can show a real-life example from a meme. It says, "John Cena surprises 7-year-old boy with cancer on his birthday." This sentence appears to be the title of a news article. The humor many people see from the meme is due to the fact that the expression can be interpreted as John Cena giving the boy cancer on his birthday. But what the article likely meant was John Cena surprised the boy for his birthday by showing up to the hospital. One sentence appears to have two different meanings. The ambiguity in the sentence can be explained visually with two synapper models.
\begin{figure}[ht]
    \centering
    \includegraphics[width=0.75\textwidth]{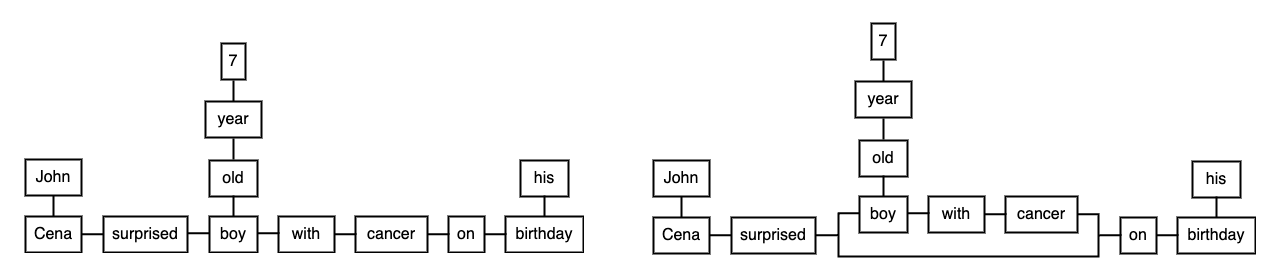}
    \caption{What appears to be the same structure linearly shows up as two distinct synapper models. (The loops connecting the beginning and the end of the sentences are removed for simplification.)}
    \label{fig:fig05}
\end{figure}
\par
The left synapper model on Figure \ref{fig:fig05} shows the humorous interpretation. The right synapper model correctly displays what looks like the intended meaning of the sentence. The difference is whether or not the phrase \emph{with cancer} is part of the embedded loop describing the boy. In the left synapper model, the phrase describes the verb \emph{surprised} and is not part of the constituent about the boy. The essence of the unintended meaning is \emph{Cena surprised him with cancer} (Cena gave the boy cancer) whereas the essence of the intended meaning is \emph{Cena surprised him who has cancer} (The boy already had cancer before meeting Cena).
\par
We should also note that a single universal syntactic structure can have more than one meaning. This is because a word can be interpreted in more than one way; or it could belong to more than one lexical category. Thus, it is possible for the same syntactic structure to defer semantically. For example, the word \emph{orange} as in \emph{Gary's choice was orange} can refer to either the fruit or the color.
\section{Case Study}
We further examine the potential effectiveness of synapper modeling for MT by putting it to test with a complex Korean sentence. Then we compare the result with currently available machine translation services such as Bing Microsoft Translator, Google Translate, and Naver Papago.
\par
The following is a sentence from a news article by Yonhap News, 불붙는 우주관광…베이조스 오는 20일 여행도 항공당국 승인 (Space travel heating up... Bezos also approved for travel on the upcoming 20th by the Federal Aviation Administration):
\begin{extract}
영국 억만장자 리처드 브랜슨 버진그룹 회장의 민간 우주 관광 시험비행이 성공하며 '스타워즈 시대'의 포문을 연 가운데 미 연방항공국(FAA)이 제프 베이조스 아마존 이사회 의장이 이끄는 블루 오리진의 유인 우주비행을 승인했다고 로이터 통신이 12일(현지시간) 보도했다 (Kim 2021).
\end{extract}
\par
We should note that a synapper model can have more than one meaning in some circumstances. If a word used in a sentence has more than one definition or if it belongs to more than one lexical category, the same structure can defer semantically. The word \emph{orange} as in \emph{Her answer was orange} can refer to a fruit or a color.
\par
This declarative sentence contains 35 words. An English translation by a human is as follows:
\begin{extract}
Reuters News Agency reported on the 12th (local time) that the U.S. Federal Aviation Administration (FAA) approved manned space travel from Blue Origin, led by Jeff Bezos, the chairman of the Amazon Board, in the midst of opening the door to the 'Star Wars era' by succeeding a test flight for civilian space travel from a British billionaire, Richard Branson, the chairman of Virgin Group.
\end{extract}
\par
The 35 words in the original text has ballooned to 65 words for the translation in English, an 85.7\% increase. This is due to a couple of factors. First, the Korean language does not use articles such as \emph{a} and \emph{the}. So articles must be added to nouns in the English translation when applicable. Second, words or phrases such as \emph{Federal Aviation Administration} in Korean are considered single units, making them essentially one word each. Third, Korean adjectives and verbs can be grouped together, which also reduces the word count.
\par
The complexity of this Korean sentence can be challenging for the current generation of machine translation software. Having a large number of words in a sentence can exponentially increase the number of translation possibilities for what MT might consider as correct. It also likely increases the chance of producing an error in the translation since the more the number of words a sentence has, the more the number of possibilities for error exists. In fact, Google Translate gave two different Korean-to-English translations for the exact same input in Korean, alternating between the two solutions when the service was accessed on different days. Here is one of the translations given by Google Translate:
\begin{extract}
Google Translate, Version 1 (49 words):
\\
British billionaire Richard Branson, chairman of the Virgin Group, opened the 'Star Wars era' with a successful private space tourism test flight, and the Federal Aviation Administration (FAA) has approved Blue Origin's manned space flight, led by Amazon Board Chairman Jeff Bezos. Reuters reported on the 12th (local time).
\end{extract}
\par
In the original sentence, the subject–\emph{Reuters News Agency}–was located toward the end. This is somewhat unusual for the Korean language since the default word order in Korean is SOV. But, because of the extremely length of the sentence, the journalist decided to put the subject at the end of the sentence with the main verb. If the algorithm used by Google Translate fails to locate the subject properly, the translation will likely result in error. In Version 1, the English translation has a different noun phrase as the subject with the word \emph{opened} as the main verb, which is also incorrect. The translation placed the subject and the main verb of the original sentence into a separate sentence.
\begin{extract}
Google Translate, Version 2 (50 words):
\\
The U.S. Federal Aviation Administration (FAA) has approved Blue Origin's manned space flight, led by Amazon Board Chairman Jeff Bezos, as British billionaire Richard Branson, chairman of the Virgin Group, successfully test flights for private space tourism, ushering in the "Star Wars era" Reuters reported on the 12th (local time).
\end{extract}
\par
Version 2 correctly translates the source as one sentence. Overall, the translation holds the essence of the original text's message. However, the words \emph{has approved} in the beginning of the sentence should simply be \emph{approved} as in \emph{approved on the 12th of July} since the news article is reporting what took place on a particular date. Also, because of the way the words are ordered, it is somewhat ambiguous whether Jeff Bezos led Blue Origin's manned space flight or that he led the U.S. Federal Aviation Administration (FAA). This confusion does not exist in the original sentence.
\begin{extract}
Bing Microsoft Translator (44 words):
\\
British billionaire Richard Branson's successful private space tourism test flight opened the door to the "Star Wars era," reuters reported on Thursday (local time) that the FEDERAL AVIATION ADMINISTRATION (FAA) had approved Blue Origin's manned space flight, led by Amazon Board Chairman Jeff Beizos.
\end{extract}
\par
Although this translation may be adequate for comprehension, it combines two different thoughts as one in the form of \emph{A opened B, it reported that X had approved Y}. This might be due to the fact that the MT algorithm could not decipher what was actually reported by Reuters while still requiring the translation to be a single sentence. In addition, the word \emph{Thursday} is not present in the Korean sentence but was added to the English translation somehow. The date mentioned in the news article is supposed to be July 12, 2021, a Monday.
\begin{extract}
Naver Papago (42 words):
\\
The Federal Aviation Administration (FAA) has approved a manned space flight of the Blue Origin, led by Amazon Chairman Jeff Bezos, amid a successful private space tourism test flight by Virgin Group Chairman Richard Branson, Reuters reported on the 12th (local time).
\end{extract}
\par
Papago is a translation service from Naver Corporation, a company based in South Korea. The translation result is somewhat similar to Google Translate's (Version 2) in terms of its structure. However, it is missing an entire segment of the original text regarding the Star Wars era.
\par
Synapper modeling of the same sentence takes a completely different approach. Here we shall address the fact that it does not technically translate sentences from one language to another in the traditional sense. Instead, synapper modeling constructs the correct syntactic structure of a sentence for all languages (language-independent) and then produces output in the targeted language (language-dependent). Since English's word order is SVO and Korean's word order is SOV, the words of the synapper model for the original sentence have to be read in the opposite order for English. However, because the journalist put the subject at the end of the sentence, it is no longer an SOV sentence. So the subject has to be moved to the beginning of the sentence to make the sentence's word order SOV.
\clearpage
\begin{figure}[ht]
    \centering
    \includegraphics[width=0.75\textwidth]{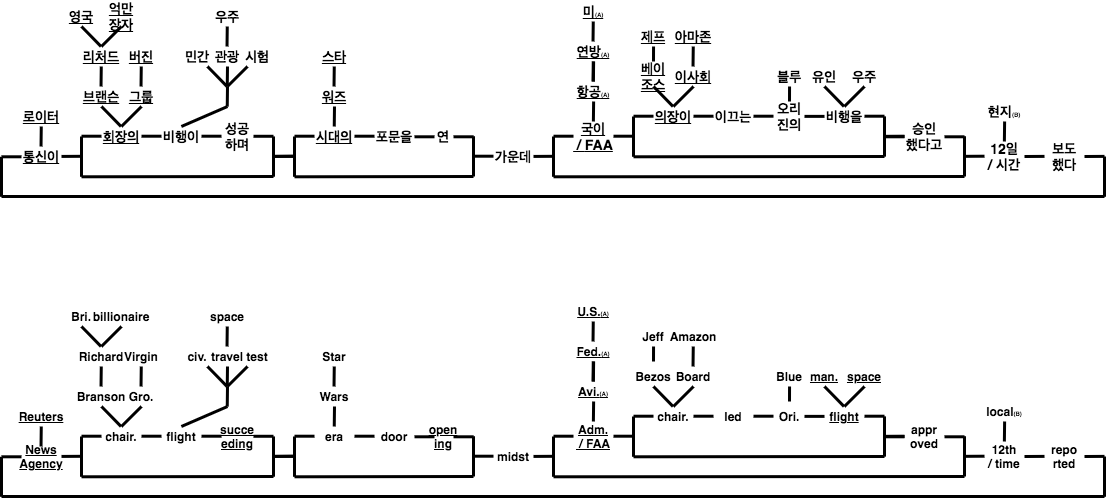}
    \caption{The syntactic structures are 100\% identical for the two languages.}
    \label{fig:fig06}
\end{figure}
\par
Once the words are changed from Korean to English, the synapper model can generate the correct English translation. To make the sentence SVO as in English, it starts with the subject noun phrase followed by the main verb phrase and then finishes with the rest of the sentence by traveling counterclockwise for all the loops present in the model. The following is the outcome:
\begin{extract}
Reuters News Agency reported \textbf{on the} 12th (local time) \textbf{that the} U.S. Federal Aviation Administration (FAA) approved \textbf{a} manned space flight \textbf{from} Blue Origin led \textbf{by} Jeff Bezos, \textbf{the} Amazon Board chairman, \textbf{in the} midst \textbf{of} opening \textbf{the} door \textbf{to the} Star Wars era \textbf{by} succeeding \textbf{a} civilian space travel test flight \textbf{from a} British billionaire Richard Branson, \textbf{the} Virgin Group chairman.
\end{extract}
\par
43 words were derived from the synapper model. When articles and prepositions are added (as shown in bold), the total number of words in the sentence increases to 62, which nearly matches the 65 words in the human translation. Also, the output does not have any of the inaccuracies that were discussed in the five translation results from the four web services. This is likely due to using no probabilistic computations, which would cleave the sentence into parts and reassemble them for the output. By keeping the syntactic structure intact, any nuance or human element present in the source is much more likely to remain in translation.
\section{Further Analysis}
Bilingual Evaluation Understudy–commonly known as BLEU–is a metric designed for evaluating or comparing the quality of machine translation results. It measures how many of the words in the machine translation matches that of the human translation of a particular sentence. For instance, if the machine translation generates \emph{I'm happy to hear the news} from a sentence in another language, whereas the human translation of the same sentence is \emph{I'm thrilled to hear the news}, then the BLEU score would be 5/6 (0.83) since only one out of the six words is different between the two translations. The best BLEU score for any translation is 1. Although BLEU is not supposed to be a comprehensive evaluation method of machine translation quality, it is quite popular and easy to use. Many NLP researchers find it helpful for comparing two different MT algorithms. A typical BLEU score can vary greatly depending on what the source and target languages are and what corpus is being used. In one study, English translations of Chinese sentences by machine translation services such as Google Translate and DeepL received BLEU scores between 0.1934 and 0.2284 (Liu and Zhu 2023). Instead of comparing these scores to the perfect score of 1, looking at the differences between the result of each translation service provides more insight on the quality of machine translation. However, when a completely different approach of MT is taken, metrics such as BLEU may not be as useful and could be even considered as inappropriate use (Callison-Burch, Osborne, and Koehn 2006).
\par
Our initial plan was to use BLEU to compare the performance of syntactic-based translation (SBT) to the currently popular neural-based translation (NBT) approach from Korean to English. Since MT services such as Google Translate already performs quite well for sentences with 15 words or fewer, we looked to compare SBT to NBT by mainly analyzing longer sentences with higher complexity. However, we instantly ran into one major obstacle after another, making the quantitative assessment essentially impossible to even conduct. Finding a large corpus with many translated sentences was not necessarily an issue. But a close examination revealed most of the translated sentences in English lacked enough quality to be set as the standard for any kind of test. In all likelihood, we suspect they were either created by non-professional translators or even generated by machine translation services. Comparing MT performances with a heavily flawed dataset would be no use. If we were to compare weather forecasts generated by a supercomputer, the data should be analyzed in relation to the actual weather data collected from the real world. If the real-world weather data is unreliable or corrupt, then no meaningful analysis about the performance of the weather forecasts can be drawn.
\par
Even some of the sampled Korean sentences turned out to be problematic for a quantitative analysis. All of the pre-translated sentences were either written or spoken by humans and for humans. This allows a certain degree of grammatical freedom for the corpus. One example would be subject omission. When the reader or listener is expected to know the subject, either a noun phrase or a pronoun may be implied. Another example is compound sentences, where two independent sentences are conjoined by a conjunction. Many reference sentences did not separate words properly or made punctuation errors as they often take place in writing. Perhaps due to these errors, the translation would sometimes take the liberty to turn one Korean sentence into two English sentences. When a sentence becomes long and translation becomes somewhat challenging, the result usually became a rough translation at best. In theory, the BLEU metric may provide meaning information about translation quality. However, this does not seem to be the case in practice. It may not make much sense to use metrics such as BLEU without having a gold-standard human evaluation, which may not be feasible especially with a large corpus and multiple domains (Reiter 2018).
\par
For this reason, we adjusted our approach to analyze shorter sentences instead. The intention was to find well-written reference sentences that did not contain grammatical errors. This would, in turn, make their translations more precise as well. We randomly sampled 64 reference sentences in Korean in five different domains from a Korean-English translation corpus on AI-Hub. On average, the length of an original sentence in Korean is 10.45 words and the length of a translated sentence in English is 19.03 words. Korean sentences usually have shorter word counts compared to English sentences, so the difference of length is not atypical. However, only 20 out of the 64 translations turned out to have the exact same meanings as the reference sentences. The other 44 translations were insufficient to be considered as acceptable translations. For instance, one sentence, \emph{뒤를 보니 어느 아주머니가 흰옷을 입은 채 눈을 치켜뜨고 죽어 있었다}, should have been translated as \emph{When I looked behind me, some middle-aged woman lay dead with her eyes looking up while being dressed in white clothes}. But its reference translation in the corpus is \emph{Looking back, a lady was lifting her eyes in white clothes and dying}. The Korean sentence clearly indicates the subject is deceased. However, she is described as alive in the reference. About 69\% of the translated sentences from the sample contained similar errors that ended up changing the meaning of the original sentence to some extent. The poor quality of translation is likely due to the translators lacking proficiency or qualification. Also, the high demand for generating big data required for training models puts an enormous emphasis on the data's quantity and not on its quality. Only upon a very painstaking and time-consuming analysis by highly qualified translators in each domain of translation one can realize significant shortcomings of the translated sentences. Otherwise, the marginal importance of using metrics to compare different MT models may not get noticed by anyone. Therefore, obtaining a BLEU score would be nothing more than a futile endeavor from this kind of data where the reference is very much flawed. Training neural-based MT models with the same corpus will most likely deteriorate the performance of translation as well. But this deterioration of translation quality may not get detected if the data in use is too large to be manually inspected by humans and metrics like BLEU also fail to provide accurate assessments. Furthermore, analyzing relatively shorter sentences diminishes a prospect of displaying any stark differences between NBT and SBT in the first place. Lengthy sentences with more complexity in their syntactic structures would be more ideal for observing the true performance of syntax-based translation. Racing at the maximum speed of 60 mph (97 km/h) will not be indicative of how much faster a sports car is compared to an average mid-sized sedan. But racing at 120 mph (193 km/h) or faster will more likely show the difference in performance between the two vehicles. Similarly, complex sentences are better for gauging SBT's performance in comparison to NBT. But the quality of reference text is likely going to decline as the average length of sentences increases. Therefore, we could not find any appropriate use of the BLEU metric and it had to be disregarded altogether. Nonetheless, we decided to continue our analysis for the 20 sentences that came with proper reference translations.
\clearpage
\begin{figure}[ht]
    \centering
    \includegraphics[width=0.75\textwidth]{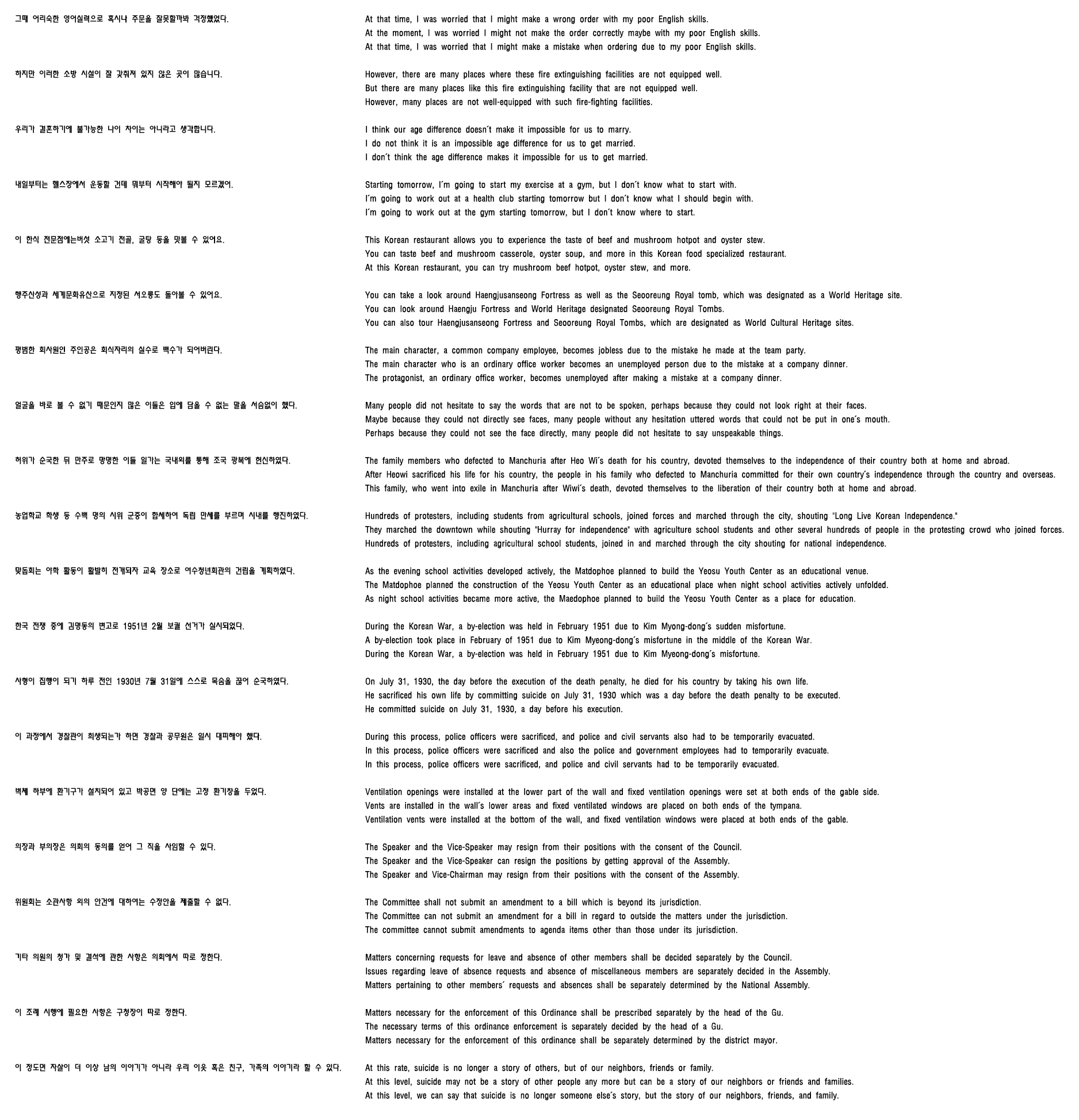}
    \caption{The translations (right column) of 20 Korean sentences (left column) in the order of reference (humans), synapper model (SBT), and Google Translate (NBT) per translation.}
    \label{fig:fig07}
\end{figure}
\par
For the new set of sampled Korean sentences, the total word count for the reference translations is 381. The synapper model translations net 391 words in total. For Google Translate, the total is 341 words for its translations. As we expected, both approaches of MT produced quite similar outcomes. All of the reference, SBT, and NBT translations are almost indistinguishable from each other in terms of translation quality. It is important to note here that the analysis mainly pertains to syntax. A lexical analysis of word conversions from Korean to English for the sample was not conducted for this article. We also attempted to translate the already translated sentences in English back into Korean. Google Translate was used for both SBT and NBT. When we compared the "reversed translations" to the original sentences, we also could not detect any significant difference between the SBT and the NBT results. This indicates the current neural-based approach like Google Translate works quite well for sentences with medium complexity. What it says about the syntax-based approach is somewhat nuanced. The issue with SBT is not the performance. If anything, the results show SBT perform at the level of NBT, if not better, for sentences in medium length. This could be considered as a major feat for two reasons. First, SBT appears to be universal for all natural languages. Second, this level of performance can be achieved without data or training, which are necessities for NBT. We may have found a way for the machine to process language that is the equivalent to how the machine handles numbers. A natural language processing system with the kind of speed and accuracy that digital computers have for making mathematical calculations could be a breakthrough. But this applies to NLP in general. For machine translation, however, SBT by itself appears to work almost a little "too well." White universal syntactic structures appear to unify two very distinct languages like English and Korean by using synapper models, the translation results seem somewhat technical, and therefore, not as natural. We believe the ultimate solution to machine translation is to use both neural-based and syntax-based approaches, where the basis of a translation is determined by the sentence's syntactic structure and then various modifications could be made to the initial translation to make it appear more human-like. Some domains will require more fine-tuning than others. For example, conversational speech texts might need to sound a lot more like humans talking to each other. But legal documents such as contracts may not have the same need to sound less technical, which might actually end up lowering the quality of translation.
\par
The objectivity of this type of analysis may be somewhat in question. Metrics like BLEU can provide some meaningful information about translation results if they can accurately analyze the syntactic structure of a sentence. However, this means the syntactic structure must be determined. In order to do so, we would need to use synapper models to figure out sentences' universal syntactic structures. But this method would only end up giving 1.0 to the BLEU score of SBT translations every time whereas the NBT's BLEU score could be significantly lower. (Once again, we are only factoring in syntax for comparison.) Even human translations in general would be scored below that of SBT. Although this could be a rather fair assessment overall, technically speaking, it still does not serve the purpose of analyzing the SBT approach objectively. Because of this reason, BLEU should not be used for testing.
\par
On a different note, we decided not to make exact calculations for measuring the probability of 20 random Korean sentences producing the correct word order for each word in every translated sentence in English as a coincidence. We estimate the odds to be somewhere around one out of a centillion, a number with one followed by 303 zeros (\begin{math} 1 \times 10^{303} \end{math}).
\par
Analyzing the syntactic structure with synapper models could turn out to be an effective method of generating quantitative data that can compare translation results in an objective manner. This is due to the fact that it would compare the syntax of two different sentences and not simply word matching. If we believe universal syntactic structures do exist for grammatical declarative sentences and semantics is indeed correlated to the sentence's syntactic structure, then it might be the only effective way to evaluate translation performances quantitatively between two different models.
\section{Interpretation}
The nature of universal syntactic structures can be viewed from several different perspectives. It can be interpreted linguistically using the three components of word order; subject, object, and verb. In graph theory, a synapper model is essentially a directed graph with vertices (nodes) and directed edges (lines). Any end-point vertex or leaf vertex is considered a branch.
\clearpage
\begin{figure}[ht]
    \centering
    \includegraphics[width=0.75\textwidth]{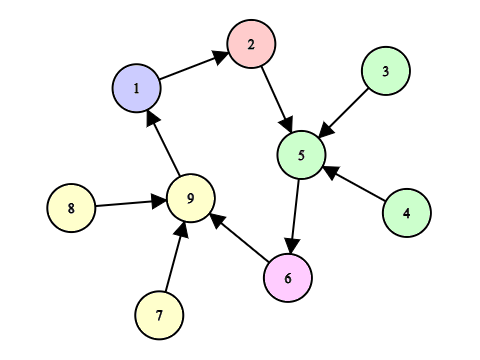}
    \caption{A synapper model as a directed graph (with matching colors for nodes with associated branches).}
    \label{fig:fig08}
\end{figure}
\par
In engineering, a synapper model can be viewed as an electric circuit. The circuit must be closed in order to form a loop. This enables electricity or information to flow from one node to another. With an open circuit, on the other hand, a current cannot flow and therefore sentence formation most likely does not occur. It is also possible to view a synapper model as a neural circuit in the brain. In either case, a closed loop is necessary for processing words one by one by allowing a current to flow.
\par
The fairly recent development of Universal Dependencies (UD) attempts to generate morphosyntactic representations of various types of sentences for different languages. By providing annotation for dependency representations of how words are dependent of each other in different sentences, it is possible to create a collection of the annotation commonly known as treebanks. This effort has been proven quite useful in NLP and many different treebanks have been developed by various groups for different languages. But the key issue is that UD has not actually reached the stage of being "universal" across all natural languages. This is the reason why so many treebanks exist because Dutch treebanks likely will not work for other languages like Turkish (de Marneffe and Nivre 2019). Also, the existence of different annotation schemes even within the same language raises the question whether UD actually shows syntactic representations or it just simply annotates syntactic-semantic relations between words. At the same time, the use of UD in NLP nevertheless has been found quite effective, since it does provide some morphosyntactic information for language processing. In other words, it gives deep neural networks something rather than nothing, practically turning unsupervised learning into supervised learning. This shows the potential effectiveness of vectors and symbolic representations conjoined for optimal performance.
\par
Tree diagrams–traditionally being used by linguists and syntacticians for representing sentence structures–fail on two grounds. One, splitting sentences into two components such as noun phrase (NP) and verb phrase (VP) does not unify the six possible word orders. Only when a sentence is split into three main components such as subject (V), object (O), and verb (V) the six word orders unify as a single structure. Secondly, syntax trees are not closed circuits and they fail to explain how a current can flow from word to word. Without an electric current, the ability to process words one by one for the brain may not be possible especially with such accuracy and speed. It appears the only conceivable way to create universal syntactic structures is to use multiple dimensions to connect words in closed loops with embedding for recursion when necessary.
\begin{figure}[ht]
    \centering
    \includegraphics[width=0.75\textwidth]{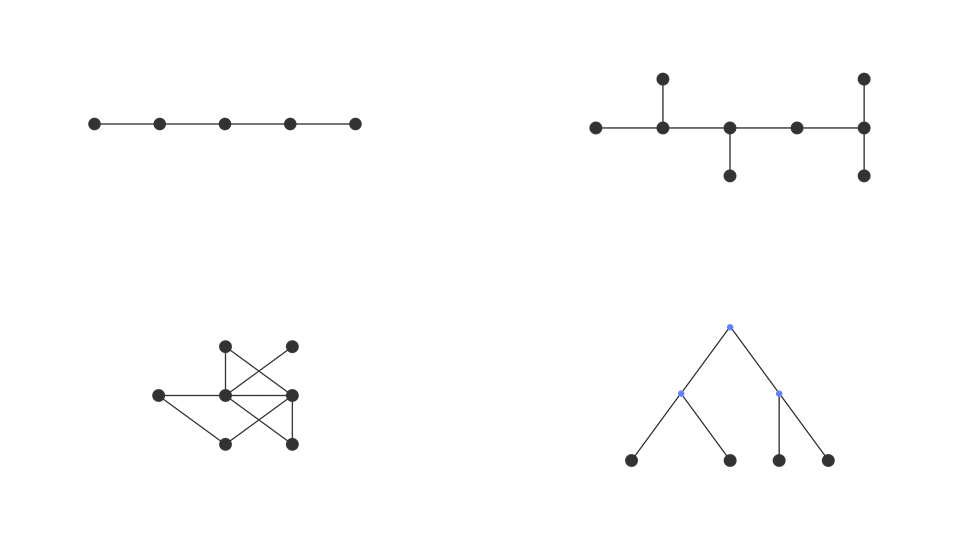}
    \caption{The number of possible ways of connecting words for sentence formation is fairly small. Only one of them (top right) appears to produce universal syntactic structures when looped.}
    \label{fig:fig09}
\end{figure}
\par
Therefore, we are left with no choice but to consider the possibility of perhaps having discovered the human brain's mechanism of language processing. Otherwise, there is no other scientific explanation of our ability to utilize natural languages and translate one natural language to another. With the observation of synapper models producing human-level translation quality in our research, we believe this is surprisingly consistent with the production of language utilization by the human brain. The results may not be necessarily conclusive at the given time, but they nonetheless appear to be promising enough to entail a more in-depth look at the phenomenon.
\section{Universal Grammar}
In late 1970, a social worker at a welfare office noticed a strange-looking girl with her mother. The girl's story later turned into a famous documented case described by the press and the scientists as Genie. She is the daughter of Clark and Irene Wiley, a victim of one of the most egregious cases of child neglect and abuse. Genie's father locked her in a small room where she would not receive almost any kind of normal human interaction or auditory stimulation for the first 13 years of her life up until she was rescued by the government. Even though she is technically not a feral child or wild child, Genie's case intrigued many linguists and psychologists since she was not exposed to language until she hit the age for puberty. Victoria Fromkin, a linguist, wanted to show with Genie that language can be acquired after puberty. Not only Genie had no prior exposure to language but she also showed signs of high intelligence despite her extremely devastating past living conditions. In some cognitive tests, the girl scored higher than any other person on record. However, no amount of effort by the scientists could teach her grammar. Genie ultimately failed to speak or sign like a normal child. Susan Curtiss–a linguist who worked intensively with Genie–reported that Genie could not make use of pronouns or WH question words like \emph{where} and \emph{how} (Rymer 1994).
\par
The use of interrogative words would require movements of words by their parts of speech. For instance, a sentence such as \emph{The toy is missing} in its declarative form must move the verb to the front of the noun phrase to form a question (e.g. \emph{Why is the toy missing}?). A pronoun such as \emph{they} can replace an entire noun phrase like \emph{the horses on the field}. Genie likely could not acquire these techniques due to not having syntax in her brain. Her language faculty did not get stimulated for the first 13 years of her life and grammar was no longer intact. Genie's case is a prime example of critical period hypothesis, an assertion that a human must be exposed to language very early in one's life in order to utilize language later on. This claim is further supported by brain scans of people suffering from language disorders. Patients like Genie do not display lateralization on the left hemisphere, which is crucial for language processing. This suggests language is an innate property of the human brain but with a time limit.
\par
The theory of universal grammar also should be reassessed. With synapper modeling, we have demonstrated the possibility of syntax-semantics unification. Sentences in different natural languages with the same meaning appear to have the same syntactic structure. If the syntactic structure of a thought is indeed identical for all natural languages, this could very well be proof of Chomsky's claims that language is innate and all natural languages are compatible with each other (Chomsky 2000). Chomsky and several other linguists have long suspected that the grammars of various languages only differ in the setting of certain innate parameters among possible variants (Carnie 2002).
\section{Language and Learning}
Most discussions on the topic of language do not involve defining the word. It is generally assumed everyone already knows what language is. Most definitions of the word in dictionaries make use of the word \emph{communication} to define it. Although language can very well be used as a "form of communication," it does not appear to be its nature; just like language is not a form of music just because it is used as lyrics for singing. What universal syntactic structures seem to reveal is language is a system of connecting words that can generate declarative sentences bounded by certain rules. When words are linked together syntactically based on their parts of speech and what they refer to, it creates a gestalt of thought. This creation is something different from a set of words put together in a random manner. It creates some kind of a structure of neural networks that potentially give rise to meaning. In other words, language is an encoding and decoding system of vocabulary and syntax for generating discrete thoughts. Despite showing high levels of intelligence on spatial tests and in her behavior, Genie never developed language. She appears to have lost her innate ability of encoding and decoding structural sentences after 13 years of receiving no lingual exposure. "What is it that language can do for a person?" Susan Curtiss answered her own question after spending years of research with Genie, "It allows us to cognize, to think, and that's important to me, because I'm that kind of a person. It also allows us to share ourselves with others–our ideas and thoughts. And that provides a huge part of what I consider to be human in my existence (Rymer 1994)."
\par
Today's large language models (LLMs) do not implement the same encoding/decoding process of language utilization, universal syntactic structures. Instead, it is mostly a game of probability where the quality of its output depends on training and the data used. For instance, LLMs have trouble with object relative clauses such as the following:
\begin{center}
The authors that the minister likes (laugh/laughs).
\end{center}
\par
The accuracy of LLMs making correct predictions such as whether the verb \emph{laugh} in this example should be in agreement with \emph{the authors} or \emph{the minister} can get as low as 18\% correct. (In this example, the object relative clause is \emph{that the minister likes}, which modifies the subject \emph{the authors}. Therefore, the correct answer is \emph{laugh} and not \emph{laughs} since it is the authors that are doing the laughing.) Surprisingly, humans were not perfect, either. The accuracy levels of the human performances in the same experiments ranged from 96\% to 72\% correct (Marvin and Linzen 2018). But analyzing the syntactic structure of this sentence clearly shows what the correct answer is. Since every word in the sentence except the verb \emph{laugh} is embedded as one unit (subject + relative clause), it is quite clear to see that \emph{laugh} should be in agreement with the subject \emph{the authors}. Without the analysis of syntactic structures, LLMs have no choice but to guess the answer. Something that is as simple and basic as guessing the correct number of sentences in a paragraph from a news article turns out to be problematic for ChatGPT. We have discovered that ChatGPT constantly gives wrong answers when asked to count the number of sentences in a given paragraph in English. Without a true understanding of how language works, LLMs fail to carry out even basic tasks that are language related. Once LLMs can analyze the syntactic structure of sentences, we can expect them to perform at the human level or perhaps even better at natural language processing. This expectation is, of course, based on the implication that universal syntactic structures allow LLMs to have the proper encoding and decoding algorithms like the human brain.
\par
It may seem that we can generate discrete thoughts by creating universal syntactic structures, but this is not technically true. We must distinguish the difference between thoughts and representations of thoughts. Genie clearly showed the ability to think like everyone else. What she could not perform was expressing and sharing her thoughts with language utilization. Hence giving knowledge to the machine in the form of universal syntactic structures will not make it intelligent. To be intelligent like humans, the machine has to decode the data the way our brains do and turn them into discrete thoughts. Without understanding how the brain thinks–whether it is a human brain or an animal brain–at the prelingual level, we may not be able to make the machine think exactly like humans. This suggests the machine also may not be able to understand concepts the same way we learn. Even with artificial neural networks that mimic biological neural networks and "sensory input devices" such as cameras and microphones that can function as artificial eyes and ears, it still will not be able to comprehend abstract concepts such as \emph{peace} and \emph{civilization}. However, we can still develop novel algorithms that can decode and analyze universal syntactic structures for the machine to be able to comprehend and think in its own way. With reverse engineering and many attempts of trial and error, language models might become better at matching human cognition while significantly reducing the necessary amounts of data and computing power.
\section{Conclusion}
The objective of this article is rather simple; it is to provide convincing evidence for the existence of universal syntactic structures and what it entails for AI. The discovery of such structures that can reveal the hidden architecture of what we call \emph{language} has been the desire for many linguists, neuroscientists, philosophers, psychologists, computer scientists, and others. Like the double-helix structure of DNA giving us insight into how genetics works, universal syntactic structures could tell us how language works. This, in turn, can potentially play a crucial role in the development of artificial general intelligence. Understanding what language is and how it works would be the first step toward building a machine that can think and reason like we do. For this reason, we request this article to be thoroughly examined and viewed as critically as possible. One should not agree with any claim made in this article without much scrutiny. In fact, it might be more constructive for one to attempt to disprove the claims with convincing evidence. Only when one fails to do so, then we can discuss the possibility of having a discourse at a large scale in order to form a consensus for a verdict. Since this has not been established yet, the majority of this article is devoted to introducing the idea of universal syntactic structures through the analysis of syntactic structures of various natural languages that appear to be identical. Because of this reason, more important aspects of this potential discovery are not fully discussed in this article. In our view, it is important that we look further into this topic much more broadly than just considering it as a new approach to machine translation or even natural language processing. The analysis of the syntactic structures of various logical sentences can decipher how the brain encodes and decodes discrete thoughts. Based on this type of research, we may be able to implement logic, reason, and meaning into the computer as algorithms that can yield the cognitive behavior human brains exhibit.
\par
The application of synapper modeling for machine translation has many advantages over today's predominant computation-driven approaches. By design, probabilistic models of machine translation such as SMT and NMT must use approximation for result (Johnson et al. 2017). Although incremental changes can be applied to improve performance, the effect of diminishing returns will eventually pervade with time. The same phenomenon can be observed in other areas such as weather forecasting and board gaming. The amount of improvement that can be obtained is almost always greater in the initial stage of development than later. This is a limitation of taking probabilistic approaches. Synapper modeling, on the other hand, gets rid of this drawback significantly. We speculate that the human brain perhaps utilizes the same basic mechanism for the utilization of language such as translation and sentence formation. If so, implementation of this system in MT will likely improve the quality of machine translation to the level of human translators without requiring considerable computing power.
\par
Perhaps this may not be such a surprising outcome considering Chomsky's long-held proposition that "linguists must be concerned with the problem of determining the fundamental underlying properties of successful grammars. The ultimate outcome of these investigations should be a theory of linguistic structure in which the descriptive devices utilized in particular grammars are presented and studied abstractly, with no specific reference to particular languages" (Chomsky 2015).
\section*{Acknowledgments}
No financial support was received for this research. No part of the content was created by any form of artificial intelligence.
\section*{References}
Bahdanau, Dzmitry, KyungHyun Cho, and Yoshua Bengio. 2015. Neural Machine Translation by Jointly Learning to Align and Translate. In \emph{Proceedings of ICLR-2015}, pages 1–2.
\smallskip
\noindent
Barman, Binoy. 2012. The Linguistic Philosophy of Noam Chomsky, pages 117–118. \emph{Philosophy and Progress}, 51(1–2), 103–122.
\smallskip
\noindent
Callison-Burch, Chris, Miles Osborne, and Philipp Koehn. 2006. Re-evaluating the role of BLEU in machine translation research. In \emph{Proceedings of EACL-2006}, pages 249–256.
\smallskip
\noindent
Carnie, Andrew. 2002. \emph{Syntax} (1st ed.), pages 18–19. Blackwell Publishing, Malden, MA.
\smallskip
\noindent
Castilho, Sheila, Joss Moorkens, Federico Gaspari, Iacer Calixto, John Tinsley, and Andy Way. 2017. Is Neural Machine Translation the New State of the Art?, page 118. \emph{The Prague Bulletin of Mathematical Linguistics}, 108(108):109–120.
\smallskip
\noindent
Chomsky, Noam. 2000. \emph{The Architecture of Language}, pages 29–73. Oxford University Press, New Delhi, India.
\smallskip
\noindent
Chomsky, Noam. 2015. \emph{Syntactic Structures}, page 11. Martino Publishing, Mansfield Centre, CT.
\smallskip
\noindent
de Marneffe, Marie-Catherine and Joakim Nivre. 2019. Dependency Grammar, pages 200 and 209. \emph{Annual Review of Linguistics}, 5:197–218.
\smallskip
\noindent
Johnson, Melvin, Mike Schuster, Quoc V. Le, Maxim Krikun, Yonghui Wu, Zhifeng Chen, Nikhil Thorat, Fernanda Viégas, Martin Wattenberg, Greg Corrado, Macduff Hughes, and Jeffrey Dean. 2017. Google's Multilingual Neural Machine Translation System: Enabling Zero-Shot Translation, page 344. \emph{Transactions of the Association for Computational Linguistics}, 5:339–351.
\smallskip
\noindent
Kim, Jin-bang. 2021, July 13. 불붙는 우주관광…베이조스 오는 20일 여행도 항공당국 승인 [Space travel heating up... Bezos also approved for travel on the upcoming 20th by the Federal Aviation Administration]. \emph{Yonhap News Agency}.
\smallskip
\noindent
Kondratyuk, Dan and Milan Straka. 2019. 75 Languages, 1 Model: Parsing Universal Dependencies Universally, page 9. In \emph{Proceedings of EMNLP-IJCNLP-2019}, pages 2779–2795.
\smallskip
\noindent
Liu, Shanshan and Wenxiao Zhu. 2023. An analysis of the evaluation of the translation quality of neural machine translation application systems. \emph{Applied Artificial Intelligence}, 37:1.
\smallskip
\noindent
Marvin, Rebecca and Tal Linzen. 2018. Targeted Syntactic Evaluation of Language Models. In \emph{Proceedings of the EMNLP-2018}, page 8.
\smallskip
\noindent
Reiter, Ehud. 2018. A structured review of the validity of BLEU. 2018. \emph{Computational Linguistics}, 44(3):393–401.
\smallskip
\noindent
Rymer, Russ. 1994. \emph{Genie}, pages 9, 17–18, 123, 128, and 220. Harper Perennial, New York, NY.
\end{CJK}
\end{document}